\documentclass{article}

\usepackage[main, preprint]{template}

\usepackage[utf8]{inputenc} % allow utf-8 input
\usepackage[T1]{fontenc}    % use 8-bit T1 fonts
\usepackage{hyperref}       % hyperlinks
\usepackage{url}            % simple URL typesetting
\usepackage{booktabs}       % professional-quality tables
\usepackage{amsfonts}       % blackboard math symbols
\usepackage{nicefrac}       % compact symbols for 1/2, etc.
\usepackage{microtype}      % microtypography
\usepackage{xcolor}         % colors
\usepackage{graphicx}
\usepackage{amsmath}
\usepackage{multirow}
\usepackage{array}
\usepackage{tcolorbox}
\usepackage{makecell}
\usepackage{caption}
\usepackage{subcaption}
\usepackage{tabularx}
\tcbuselibrary{skins, breakable}
\usepackage{cleveref}
\usepackage{enumitem}

\newtcolorbox[auto counter, number within=section]{infobox}[2][]{
    colback=gray!10!white,
    colframe=gray!70!black,
    colbacktitle=gray!70!black,
    coltitle=white,
    title={\textbf{\thetcbcounter: #2}},
    arc=4mm,
    boxrule=0.5mm,
    #1
}

\setlist[itemize]{
    leftmargin=*,
    itemsep=0pt,
    parsep=0pt,
    topsep=0pt
}

\title{SCPRM: A Schema-aware Cumulative Process Reward Model for Knowledge Graph Question Answering}

\author{%
  Jiujiu Chen \\
  HKUST (GZ)\\
  \And
  Yazheng Liu \\
  HKUST (GZ)\\
  \And
  Sihong Xie \\
  HKUST (GZ)\\
  \And
  Hui Xiong \\
  HKUST (GZ)\\
}

\begin{document}

\maketitle

\begin{abstract}
    Large language models excel at complex reasoning, yet evaluating their intermediate steps remains challenging. Although process reward models provide step-wise supervision, they often suffer from a risk compensation effect, where incorrect steps are offset by later correct ones, assigning high rewards to flawed reasoning paths. This issue is further exacerbated in knowledge graph (KG) reasoning, as there may exist multiple paths between the start and end entities in the KGs, and a risky step can make the reasoning path flawed. Those limitations are problematic in risk-sensitive tasks such as medical and legal KG reasoning. To address the issues, we propose a Schema-aware Cumulative Process Reward Model (SCPRM) that evaluates reasoning paths by conditioning on the reasoning prefix , and incorporating schema distance between current reasoning step and the implicit target parsed from the query, which provides cumulative and future rewards to guide the path explorations. We further integrate SCPRM into Monte Carlo Tree Search (MCTS) as SCPRM-MCTS to conduct multi-hop reasoning on KGs for question answering (QA) tasks. Across medical and legal KGQA and CWQ, SCPRM-MCTS improves the performance of Hits@k by an average of 1.18\% over strong baselines, demonstrating more accurate and risk-sensitive reasoning evaluation.
\end{abstract}

\section{Introduction}
As Large Language Models (LLMs) evolve, they have transitioned from simple question answering to complex multi-step reasoning, largely driven by the emergence of Chain-of-Thought (CoT) prompting~\cite{Wei0SBIXCLZ22}. However, the inherently linear structure of CoT makes LLMs vulnerable to error propagation, as a single mistake can derail the entire reasoning trajectory and lead to an incorrect final answer~\cite{TurpinMPB23}. While outcome-based reward models can verify final answers~\cite{abs-2501-12948, MuHHAVKLBSW24}, they cannot distinguish whether the model arrived at the correct answer through logic or hallucination~\cite{abs-2405-15092, LiC0XLJLZ25, xu2025logicreward}. To mitigate this, researchers have shifted towards Process Reward Models (PRMs) that evaluate intermediate steps combined with search algorithm, e.g., Tree-of-Thoughts~\cite{YaoYZS00N23}.

\begin{figure}
    \centering
    \includegraphics[width=\linewidth]{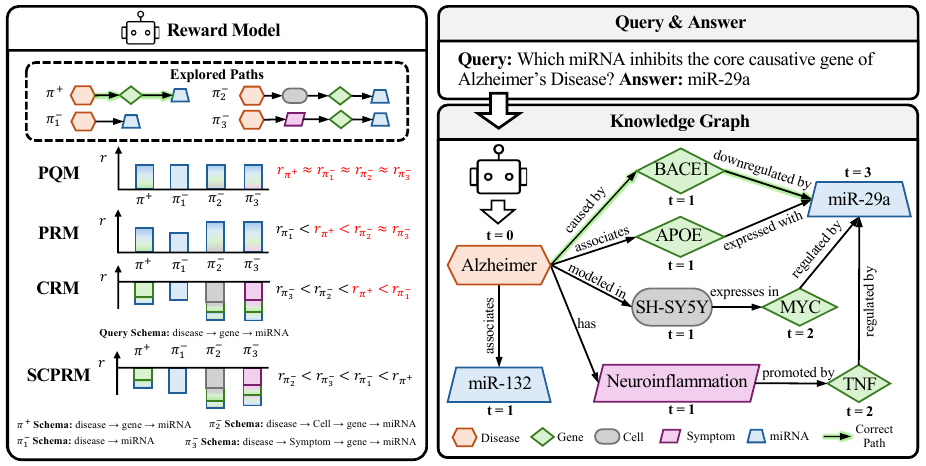}
    \caption{Comparison of reward models for reasoning based on medical KGQA. Red rewards indicate the wrong ranks of traversed paths on the KG.}
    \label{fig:fig1}
\end{figure}

Currently, existing PRMs have shown strong performance in mathematical reasoning and code generation tasks~\cite{ZhangZHYD024, LiDL0WT025}, but those tasks are typically tolerant of minor intermediate risky steps, as later correct steps can recover the final answer. Consequently, vanilla PRMs usually aggregate rewards additively across the trajectory, making them relatively insensitive to localized biases or risky decisions. This assumption breaks down in Knowledge Graph Question Answering (KGQA). The complex KG topology allows for multiple reasoning paths between the start entity and the answer, but the logically valid paths are strictly limited. This is particularly severe in risk-sensitive applications such as medical or legal KGQA (illustrated in Figure~\ref{fig:fig1}). For example, a path traversing entities related to symptoms involve logically irrelevant or incorrect information to the query, but still coincidentally reach the correct final answer. Vanilla PRMs assign such paths with high rewards by averaging risk and correctness, allowing risky steps to be offset by later rewards. In risk-sensitive reasoning tasks, such compensation is unacceptable. Once a high-risk step occurs, the remaining trajectory should be substantially penalized, even if it reaches a technically correct endpoint~\cite{abs-2509-26578}.

Beyond risk sensitivity, KGQA requires adherence to implicit constraints expressed in the query~\cite{BuCCDW0025, LiLGCZ25}. While conditional reward models~\cite{abs-2509-26578} can prevent additive reward compensation, they introduce a severe length bias. Multiplying step-wise probabilities ranging from 0 to 1 monotonically decreases the total reward, so the conditional reward models naturally tie the total reward to trajectory length, disproportionately penalizing longer reasoning path. This length sensitivity severely weakens logical exploration in KGQA, where multi-hop paths are necessary for complex queries.

To address these challenges, we propose the Schema-aware Cumulative Process Reward Model (SCPRM), designed specifically for risk-sensitive multi-hop reasoning in KGQA. Inspired by the $A^*$ search algorithm~\cite{hart1968formal}, SCPRM decouples reasoning evaluation into two dimensions: cumulative reward and future reward. Unlike prior $A^*$-style methods that rely on arbitrary heuristics or hand-crafted rule-based evaluators, SCPRM is a parameterized model with base LLM and heads for reward outputs, which is different from the design of next-token prediction. The cumulative reward utilizes step-wise conditional probabilities to capture risk propagation, ensuring that a highly risky step mathematically results in a lower reward since the step.  Simultaneously, the future reward is computed by a schema-aware estimator, which extracts reasoning logic from the query to measure the schema distance between the current step and the target, and ensures logically consistence in multi-hop paths without length bias. We evaluate SCPRM by integrating it with MCTS on KGQA tasks, and the contributions are summarized as follows:
\begin{itemize}
\item \textbf{Non-Compensatory PRM Architecture:} We introduce SCPRM, which leverages conditional probabilities to model cumulative path rewards reversed by the past cost in $A^*$ algorithm. This ensures risky steps impose persistent and multiplicative penalties, effectively vetoing risky trajectories that vanilla PRMs would artificially rescue.
\item \textbf{Schema-Aware Future Estimation:} We propose a learnable future reward module that utilize logical schema  parsed from the query. By computing schema distance between the query constraints and current reasoning, SCPRM learns the query logic by overcoming the inherent length bias of conditional reward models.
\item \textbf{Strong Performance on Risk-Sensitive KGQA:} Extensive experiments on medical and legal KGQA and Complex WebQuestions (CWQ) datasets demonstrate SCPRM with different small base LLMs (e.g., Llama3.1-8B) achieves average 1.18 \% improvement in Hits@k compared to closed-source powerful LLMs like GPT-4o-mini.
\end{itemize}

\section{Related Work}
\textbf{Reward models} are learned evaluators to score candidate outputs according to human or task preferences~\cite{Ouyang0JAWMZASR22}, and play an important role in alignment and reasoning optimization. Early reward functions were largely heuristic or rule-based, relying on manually designed criteria such as formatting constraints or keyword matching, which limited their applicability to open-ended generation~\cite{abs-2507-17746, abs-2508-12790}. Subsequent work introduced Outcome Reward Models (ORMs) that only supervise the final answer of the reasoning trajectory~\cite{abs-2110-14168}. Later studies also connected outcome supervision to value estimation for decoding, demonstrating that final-answer rewards can guide search toward more promising reasoning trajectories~\cite{YuGW24}. However, ORMs cannot localize intermediate reasoning errors due to the sparse feedback. To address this limitation, recent work has shifted toward Process Reward Models (PRMs), which assign rewards to intermediate reasoning steps rather than only final outputs. Process supervision was shown to reduce reasoning errors and outperform outcome-only supervision on challenging math benchmarks~\cite{abs-2211-14275, LightmanKBEBLLS24}. Subsequent methods improved scalability by replacing expensive human step annotations with automatically generated supervision, e.g., Monte Carlo rollouts or ranking-based Q-value formulations that model dependencies across steps~\cite{WangLSXDLCWS24, abs-2510-08049, LiL25a}. Beyond mathematics, PRMs have been extended to code generation using execution feedback as stepwise supervision~\cite{Ni0RSYWL23, BollackerEPST08, LiDL0WT025}, and medical and legal reasoning~\cite{wang2025process, abs-2506-11474, jiang2025meds, abs-2510-10072}. However, in risk-sensitive domains such as medicine and law, existing PRMs often suffer from risk compensation, where high rewards on some steps can offset critical errors. This limitation motivates the need for new process reward models that explicitly account for stepwise risk and logical safety.

\textbf{Knowledge graph question answering (KGQA)} aims to answer natural language questions by retrieving and reasoning over structured knowledge graphs, often requiring multi-hop inference across multiple entities and relations. Early KGQA methods mainly relied on embedding-based approaches that learned vector representations of entities and relations, enabling efficient link prediction and simple relational reasoning, but offering limited support for complex multi-hop decision processes~\cite{BordesUGWY13, SunDNT19}. Later work reformulated KGQA as a sequential search problem using reinforcement learning (RL), where agents learn to navigate the graph hop by hop toward answer entities. DeepPath~\cite{XiongHW17} and MINERVA~\cite{DasDZVDKSM18} improved multi-hop reasoning by learning path-selection policies, while subsequent work introduced reward shaping~\cite{LinSX18} and adversarial training~\cite{Cui0HH023} to alleviate sparse rewards and spurious paths. Recently, the integration of LLMs with KGs has catalyzed a paradigm shift, combining parametric semantic understanding with structured and verifiable knowledge~\cite{MaCWKW25}. Initial zero-shot approaches augmented LLM prompts with retrieved KG triplets~\cite{abs-2306-04136} or dense subgraphs~\cite{MavromatisK25}, though their efficacy was often bottlenecked by heuristic retrieval limitations. To achieve deeper integration, frameworks such as ToG~\cite{sun2023think}, RoG~\cite{LuoLHP24}, and PoG~\cite{TanWLXYZ25} introduced iterative exploration and path-guided reasoning, utilizing graph structures to enforce structural faithfulness. To bridge the semantic gap between complex queries and rigid graphs, recent methods leverage LLMs to dynamically align query semantics~\cite{LiLGCZ25} or guide sophisticated search algorithms like Monte Carlo Tree Search~\cite{wang2025damr} to balance exploration and exploitation. However, guaranteeing reasoning faithfulness and process-level quality without hallucination remains a critical, open challenge in multi-hop scenarios.

\section{Preliminary}
\subsection{$A^*$ Algorithm}
\label{subsec: a_star}
The classical $A^*$ algorithm~\cite{hart1968formal} is a cornerstone of heuristic search, designed to efficiently identify optimal trajectory by minimizing the cost function $f(n)=g(n)+h(n)$, where $n$ denotes the current state, $g(n)$ is the past cost and $h(n)$ estimates the future cost. 

We instead reverse the cost perspective and formulate $A^*$ as a reward-maximization problem. Our reward function is defined as $F(n)=G(n)+H(n)$, where $G(n)$ denotes cumulative past rewards and $H(n)$ indicates future rewards. This change is more than a reparameterization: it explicitly encourages high-reward intermediate steps and aligns naturally with reward-based learning. This reward-driven variant provides a simple and effective bridge between heuristic search and modern learning systems, particularly in settings where reasoning progress is better modeled by incremental gains than by minimizing cost. The implementations of $G(n)$ and $H(n)$ will be demonstrated in details in Section~\ref{sec:methodology}. 

\subsection{Problem Formulation}
Given a knowledge graph $\mathcal{G}$, the reasoning progress starts from the anchor entity $e_0$ parsed from the query $q$, and traverses $\mathcal{G}$ to obtain an optimal reasoning trajectory $\pi^{(T)}=\{(e_0, r_0, e_1), ..., (e_{T-1}, r_{T-1}, e_T)\}$, where $(e_{i-1}, r_{i-1}, e_i)$ denotes the entity triplets retrieved from $\mathcal{G}$, and $T$ denotes the trajectory steps. The goal is to traverse $\mathcal{G}$ to obtain the answer $e_T$ for $q$ in the correct reasoning trajectory. We model the multi-step reasoning on $\mathcal{G}$ as a finite-horizon Markov Decision Process (MDP) $\mathcal{M}=(\mathcal{S},\mathcal{A},\mathcal{P}, F, \gamma)$. $\mathcal{S}$ is the state space, $\mathcal{A}$ denotes the action space, $\mathcal{P}$ represents the state transition probability, $F$ is the reward function defined in~\ref{subsec: a_star}, and $\gamma \in (0, 1)$ is the discount factor. A state $s_k$ is determined by its last state and action $\mathcal{P}(s_k|s_{k-1}, a_{k-1})$. Given the state $s_{k-1}$, the actions $a_{k-1} = \mathcal{A}(s_{k-1})$ consists of all possible triplets that start with the entity $e_{k-1}$. The corresponding reasoning trajectory in state $s_k$ is represented as $\pi^{(k)}=\{ (e_0, r_0, e_1), ... ,(e_{k-1}, r_{k-1}, e_k) \}$. The trajectory probability is computed as:
\begin{equation}
    \mathcal{P}(\pi) = \prod_{k=1}^T \mathcal{P}(L(s_k) = \text{safe} | s_{k-1}, a_{k-1}) \cdot \mathcal{P} (\text{target} | s_T).
\end{equation}
The trajectory probability equivalently defines as the $A^*$-style reward computation: $F(s_T) = \log \mathcal{P} (\pi)$, which admits a decomposition: $F(s_k) = G(s_k) + H(s_k)$, where $G(s_k) = \sum_{t=1}^k \log \mathcal{P}(L(s_k) = \text{safe} | s_{k-1}, a_{k-1})$ and $L(\cdot)$ is the label indicates whether the state is safe or not. $H(s_k) = \log \mathcal{P} (\text{target} | s_k)$, and the "target" denotes the reasoning target.

\subsection{Monte Carlo Tree Search}
Monte Carlo Tree Search is a heuristic search algorithm for decision-making processes, widely adopted in domains requiring lookahead search with large state spaces. Formally, MCTS builds a search tree $\mathcal{T}$ where each node in step $k$ represents a state $s_k$ and each edge represents an action $a_k$. Each node stores a visit count $N(s_k,a_k)$ and its mean value $V(s_k,a_k)$ when given an action $a_k$. MCTS process iterates through selection, expansion, evaluation and backpropagation phases. In the selection phase, the algorithm traverses the tree by selecting an action that maximize the upper confidence bound (UCB) metric at each step:

\begin{equation}
    a_k = \arg \max_{a_i} \left[V(s_i,a_i) + c \cdot \sqrt{\frac{\ln N(s_i)}{N(s_i,a_i)}}\right],
\end{equation}
where $c \in \mathbb{R}$ is the exploration constant, which balances the exploitation guided by the value function $V(s_i,a_i)$ and exploration $\sqrt{\frac{\ln N(s_i)}{N(s
_i,a_i)}}$ terms. $N(s_i)$ is the visit count of state $s_i$. In our research, we leverage the $A^*$-style reward function as the value function of MCTS to evaluate each action $a_i$ based on current state $s_i$, and select the corresponding triplet to expand the reasoning trajectory. The state node value represents the reasoning trajectory that starts from the anchor entity to current entity, and will backpropagate to the the initial state. The final answers can be obtained from the reasoning trajectories with highest rewards. 

\section{Methodology}
\label{sec:methodology}
\begin{figure*}[htbp]
    \centering
    \includegraphics[width=\linewidth]{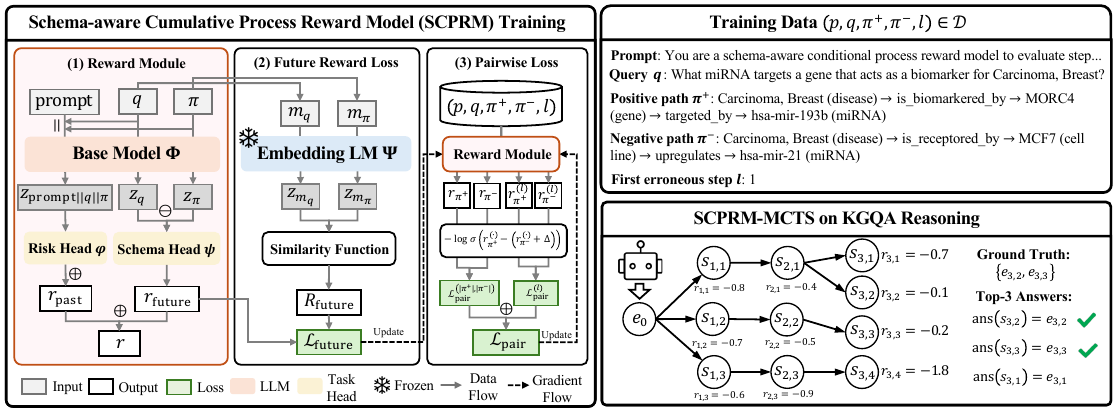}
    \caption{The full pipeline of SCPRM training and working with MCTS for reasoning on KGQA.}
    \label{fig:fig2}
\end{figure*}

\subsection{Cumulative Past Reward Modeling}
Given a query $q$ and a state $s_k$ of reasoning trajectory $\pi$, we model the step-wise safety probability as:

\begin{equation}
    p_k = \sigma (\varphi (z_{\text{prompt}||q||\pi^{(k)}})), \quad \mathcal{P}(L(s_k) = \text{safe} | s_{k-1}, a_{k-1}) = 1-p_k
\end{equation}

where the safety probability $p_k \in (0, 1)$, $\varphi$ is the risk head, and $\sigma(\cdot)$ is the sigmoid function. Hidden representation $z_{\text{prompt}||q||\pi^{(k)}} = \Phi (\text{prompt} || q || \pi^{(k)}) \in \mathbb {R}^{1 \times d}$ represents the last token of the last layer output of the base LLM, which contains the information of reasoning trajectory from the anchor entity to the current entity. The prompt forces the model to focus on the potential risks in each reasoning trajectory. $d$ denotes the dimension of hidden representation, $\Phi$ is the parameters of the base LLM, and $||$ represents the textual concatenation. 

For a given reasoning trajectory, the cumulative past reward at each state is conditioned on its prefix context. Therefore, we model the past reward by aggregating the step-wise safety probabilities along the trajectory. However, direct multiplicative accumulation leads to vanishing reward magnitudes, which can impede optimization. To mitigate the issue, we apply the logarithmic operation on the accumulation of probabilities. Consequently, the cumulative past reward is transformed to the logarithmic summation of conditional probabilities:

\begin{equation}
    G(s_k) = r_\text{past}(\pi^{(k)}) = \log \prod_{t=1}^k (1-p_t) = \sum_{t=1}^k \log (1-p_t).
\end{equation}

This formulation naturally induces an additive structure over the trajectory-level reward. Specifically, the cumulative past reward satisfies $G(s_{k+1}) = G(s_k) + c(s_k, s_{k+1})$, and $c(s_k, s_{k+1}) = \log(1 - p_{k+1})$. This additive decomposability enables step-wise accumulation of rewards and naturally supports search-based inference algorithms. Besides, it facilitates fine-grained credit assignment by attributing the overall trajectory quality to individual reasoning steps.

\subsection{Schema-aware Future Reward Modeling}
Estimating future rewards of reasoning over KG is inherently challenging in the absence of explicit targets, and existing approaches predominantly rely on semantic similarity of query and reasoning path. We propose to approximate future rewards by leveraging observed reasoning trajectories and quantifying the alignment gap between the extracted query schema and the induced reasoning schema. The schema-aware future success probability is defined as: 

\begin{equation}
    w_k = \sigma(\psi(z_q - z_\pi^{(k)})),
\end{equation}

where $\psi$ is the schema head, $z_q=\Phi(q) \in \mathbb {R}^{1 \times d}$, and $z^{(k)}_{\pi}=\Phi(\pi^{(k)}) \in \mathbb {R}^{1 \times d}$. The "target" is implicit in $q$, and $z_q$ represents the implicit target embedding. The future reward is the logarithmic probability $H(s_k) = r_\text{future}(\pi^{(k)}) = \log w_k$.

Reasoning for KGQA needs ground the knowledge provided by the KG to avoid suspicious paths. In entity exploration, we can identify each entity with its type and relational directions, which consist of the reasoning schema $m_{\pi}^{(T)}=(\text{type}(e_0) \to \text{type}(e_1) \to ... \to \text{type}(e_T))$. For the query $q$, its query schema $m_q=(\text{type}(e_0) \to \text{constraint}_1 \to ... \to \text{constraint}_K$ can be extracted by few-shot prompting any powerful LLMs with human review, and $\text{constraint}_{1...K}$ are some keywords in the initial query. For the sample in Figure~\ref{fig:fig2}, a simple query schema can be parsed as (type(Carcinoma, Breast)=disease $\to$ gene $\to$ miRNA). By comparing the reasoning schema with the query schema, we can figure out remaining information has not been fetched in the reasoning path, which consists of the "ground truth"  future reward  $R^{(k)}_{\text{future}} = \exp[-|| \Psi^{*}(m_q) - \Psi^{*}(m_{\pi}^{(k)}) ||_2^2],$. $\Psi^{*}$ denotes the parameters of the frozen embedding model that has similar architecture of $\Phi$. We utilize $R_{\text{future}}^{(k)}$ as the optimization goal for future reward computed by schema head. Therefore, the future reward loss is the mean squared error between the predicted future success probability $w_k$ and $R_{\text{future}}^{(|\pi|)}$:
\begin{equation}
    \mathcal{L}_{\text{future}} = \sum_{\pi \in \mathcal{D}} (w_\pi^{(k)} - R_{\text{future}}^{(|\pi|)})^2.
\end{equation}

\subsection{Reward Module Training}
Once we obtain the cumulative past reward and future reward, we combine them as the full reward:

\begin{equation}
    r(\pi^{(k)}) = \sum_{t=1}^k \log(1-p_t) + \log w_k \Longleftrightarrow F(s_k) = \log [ \prod_{t=1}^{k} (1-p_t) \cdot w_k ].
\end{equation}

In KGQA tasks, a query can have several valid reasoning paths leading to the correct answer, and there may also be multiple correct answer. The semantics of different feasible paths can vary, and optimizing each path’s reward independently can impede the reward model's generalization ability. In fact, the relative ranks of path rewards are more crucial than those absolute rewards. We adopt a pairwise optimization strategy for path rewards. Concretely, reasoning paths with correct answers that do not contain risky or erroneous steps are filtered as positive samples, whereas those that include erroneous steps are designated as negative samples. For pairs of positive and negative paths, we compute the difference between their rewards: 
\begin{equation}
    \label{eq:all_steps}
    \mathcal{L}^{(|\pi^+|,|\pi^-|)}_{\text{pair}} = -\sum_{(\pi^+,\pi^-) \in \mathcal{D}} \log[\sigma(r_{\pi^+} - (r_{\pi^-} + \Delta))],
\end{equation}
where $\Delta$ is the hyperparameter to increase the reward difference between the positive and negative paths. The pairwise ranking consistency is sufficient for the optimal path recovery. 

Although ranking full path rewards is important, we also need discriminate the first risky or erroneous step $l$ in the negative path to prune the low valuable paths in advance. Therefore, we also maximize the reward difference between positive and negative paths at step $l$:.
\begin{equation}
    \label{eq:part_steps}
    \mathcal{L}^{(l)}_{\text{pair}} = -\sum_{(\pi^+,\pi^-) \in \mathcal{D}}\log[\sigma(r_{\pi^+}^{(l)} - (r_{\pi^-}^{(l)} + \Delta))].
\end{equation}
 The pairwise loss is integrated as $\mathcal{L}_{\text{pair}} = \mathcal{L}^{(|\pi^+|,|\pi^-|)}_{\text{pair}} + \mathcal{L}^{(l)}_{\text{pair}}$, which ensures that SCPRM can not only penalize the paths containing risky steps without compensatory path rewards, but also discriminate erroneous steps by lower corresponding step-wise rewards. 

The total loss of training schema-aware cumulative process reward model is the summation of pairwise and future losses, with a hyperparameter $\lambda$ to control their loss contributions.

\begin{equation}
    \label{eq:total_loss}
    \mathcal{L}= \mathcal{L}_{\text{pair}} + \lambda \cdot \mathcal{L}_{\text{future}}.
\end{equation}

\subsection{MCTS-guided KG Reasoning}
Once SCPRM is trained, we apply it as the value function of MCTS to guide the path search on KGs. We start MCTS from the anchor entity $e_0$ of each query, and expand it into reasoning trajectory by concatenating with relations and neighboring entities. Each path is stepwise evaluated by SCPRM, and the reward is stored in the last tree node along the reasoning path and backpropagates through the path to the start entity. To avoid repeated exploration on the same path, we add an extra memory module to store the explored paths. Once reaching the search budget, we choose the paths of top-$k$ highest rewards, and the last entity of each path is considered as the predicted answer. 

\section{Experiments}
In this section, we aim to answer the following research questions: \textbf{RQ1}: Can SCPRM evaluate and rank trajectories more precise on risk-sensitive reasoning tasks than existing PRMs? \textbf{RQ2}: Can SCPRM-MCTS achieve better performance on risk-sensitive KG reasoning tasks than existing methods? \textbf{RQ3}: Can SCPRM-MCTS achieve comparable performance on commonsense KG reasoning tasks? 

\subsection{Experimental Settings} 

\textbf{Datasets} To analyze the performance of SCPRM designed for risk-sensitive KGQA tasks, we construct medical and legal KG and QA datasets, respectively. The medical queries involve unseen relevance between micro RNAs (miRNAs) and diseases, and the legal queries focus on crimes, legal provisions, and penalties under Chinese criminal law. The medical KG is constructed based on the data sources from HMDDv3.2~\cite{HuangSGCZLZC19} and mirTarbase~\cite{cui2025mirtarbase}, which include miRNAs, genes, diseases and PMIDs\footnote{PMID stands for PubMed Identifier. It is a unique numerical identifier to each record indexed in the PubMed database, which is the largest publicly accessible database of biomedical and life-science literature.} of relevant medical paper. We fetch the titles and abstracts of those paper via PMIDs, and extract key statements as the reference via GPT-4o-mini\footnote{https://openai.com/index/GPT-4o-mini-advancing-cost-efficient-intelligence/}. The legal KG is constructed based on CrimeKgAssitant\footnote{https://github.com/liuhuanyong/CrimeKgAssitant/blob/master/data/kg\_crime.json} and CAIL 2025\footnote{http://cail.cipsc.org.cn}, and legal concepts are considered as an entity types. We leverage GPT-4o-mini to generate queries via designed templates on the above KGs.

\begin{table}[htbp]
    \begin{minipage}[t]{0.4\linewidth}
        \small
        \vspace{0pt}
        \captionof{table}{Statistics of KGQA datasets.}
        \begin{tabular}{c|ccc}
        \toprule
        Dataset & Train & Valid & Test \\
        \midrule
        Medical KGQA & 2676 & 673 & 669 \\
        Legal KGQA & 8659 & 1562 & 2164 \\
        CWQ & 27639 & 3519 & 3531 \\
        \bottomrule
        \end{tabular}
        \label{tab:dataset}
    \end{minipage}
    \hfill
    \raisebox{-14pt}{
        \begin{minipage}[t]{0.55\linewidth}
            We also conduct experiments on CWQ dataset~\cite{TalmorB18}, which contains multi-hop queries based on Freebase~\cite{BollackerEPST08}. The statistics of datasets are demonstrated in Table~\ref{tab:dataset}, where train and validation parts are utilized for training reward models, and test part is used for evaluating the reward models on KG-based reasoning.
        \end{minipage}
    }
\end{table}

\textbf{Evaluation Metrics} We use pairwise accuracy as the evaluation metric for RQ1, and Hits@k for RQ2 and RQ3. Hit@k is computed in an answer-level manner, where a query is counted as correct if at least one ground truth answer appears in the top-$k$ predictions. Due to multiple answers to a query in medical and legal KGQA, we set $k=3$ for medical and legal evaluations. Besides, we set $k=1$ for CWQ dataset by following the settings of previous approaches~\cite{sun2023think}.\\
\textbf{Baselines} We compare our method with two categories of baselines. (1) Trainable reward models adapted for KG reasoning, including vanilla PRM~\cite{abs-2506-11474}, PQM~\cite{LiL25a} and CRM~\cite{abs-2509-26578}. (2) Prompting-based KG reasoning methods, which mainly prompt LLMs to prune and evaluate paths, including ToG~\cite{sun2023think} and ReKG-MCTS~\cite{SongZ025}. \\
\textbf{Implementations} We implement the baseline methods using their public codes, and select LLMs across different scales, including Qwen2.5-1.5B, Qwen3-4B and Llama3.1-8B. For the prompt-tuning methods, we also incorporate the GPT-4o-mini API in the experiments. The SCPRM framework comprises a trainable base LLM with reward heads, and we adopt the Low-Rank Adaptation (LoRA) strategy for parameter-efficient fine-tuning of the base LLM. The whole framework is implemented using PyTorch 2.9.1 on a workstation, which is equipped with AMD EPYC CPUs and four NVIDIA RTX A6000 GPUs, each with 48G memories. The operating system is Ubuntu 22.04 with CUDA 12.2. We use AdamW optimizer with a cosine scheduler for the learning rate, which is set $2e-4$ initially. The max number of input tokens is set 512, and the hyperparamter $\Delta$ in Eq~(\ref{eq:total_loss}) is initially set to $0.3$. For the reward model comparison, we conduct 5 folds splits on train and valid subsets. 

\subsection{Pairwise Ranking Reward Comparison}

\begin{table}[htbp] \small
    \centering
    \caption{Pairwise ranking accuracy (\%) with standard deviations of different reward models.}
    \begin{tabularx}{\linewidth}{ccccc}
    \toprule
    Methods & base LLM & Medical KGQA & Legal KGQA & CWQ \\
    \midrule
    PQM & \multirow{4}{*}{Qwen2.5-1.5B} & 72.89 $\pm$ 4.79 ($\downarrow$ 16.87) & 64.27 $\pm$ 3.82 ($\downarrow$ 19.22) & 66.38 $\pm$ 1.74 ($\downarrow$ 6.65) \\
    CRM & & 85.32 $\pm$ 1.25 ($\downarrow$ 4.44) & 80.19 $\pm$ 1.06 ($\downarrow$ 3.30) & 70.75 $\pm$ 1.36 ($\downarrow$ 2.28) \\
    PRM & & 87.71 $\pm$ 1.48 ($\downarrow$ 2.05) & 82.82 $\pm$ 0.65 ($\downarrow$ 0.67) & 71.24 $\pm$ 0.96 ($\downarrow$ 1.79) \\
    SCPRM & & 89.76 $\pm$ 1.34 & 83.49 $\pm$ 1.20 & 73.03 $\pm$ 1.05 \\
    \midrule
    PQM & \multirow{4}{*}{Qwen3-4B} & 78.22 $\pm$ 3.98 ($\downarrow$ 14.69) & 66.75 $\pm$ 2.86 ($\downarrow$ 19.68) & 70.37 $\pm$ 2.45 ($\downarrow$ 8.26) \\
    CRM & & 87.98 $\pm$ 1.37 ($\downarrow$ 4.93) & 83.54 $\pm$ 0.97 ($\downarrow$ 2.89) & 74.39 $\pm$ 1.54 ($\downarrow$ 4.24) \\
    PRM & & 89.37 $\pm$ 0.87 ($\downarrow$ 3.54) & 84.62 $\pm$ 1.12 ($\downarrow$ 1.81) & 76.58 $\pm$ 1.27 ($\downarrow$ 2.05) \\
    SCPRM & & 92.91 $\pm$ 1.17 & 86.43 $\pm$ 1.04 & 78.63 $\pm$ 1.38 \\
    \midrule
    PQM & \multirow{4}{*}{Llama3.1-8B} & 81.36 $\pm$ 3.46 ($\downarrow$ 13.98) & 70.29 $\pm$ 3.26 ($\downarrow$ 17.82) & 72.98 $\pm$ 2.33 ($\downarrow$ 8.89) \\
    CRM & & 91.54 $\pm$ 0.94 ($\downarrow$ 3.80) & 87.03 $\pm$ 1.58 ($\downarrow$ 1.08) & 79.56 $\pm$ 1.26 ($\downarrow$ 2.31) \\
    PRM & & 91.83 $\pm$ 1.91 ($\downarrow$ 3.51) & 86.70 $\pm$ 2.06 ($\downarrow$ 1.41) & 80.24 $\pm$ 1.84 ($\downarrow$ 1.63) \\
    SCPRM & & 95.34 $\pm$ 1.61 & 88.11 $\pm$ 1.47 & 81.87 $\pm$ 1.51 \\
    \bottomrule
    \end{tabularx}
    \label{tab:rm_result}
\end{table}

In this subsection, we address \textbf{RQ1} by evaluating the discriminative ranking capability of reward models. Due to the prohibitively high cost of acquiring absolute reward annotations for complex reasoning paths, we proxy this evaluation through a pairwise ranking formulation, where the correct prediction is defined as the reward model providing a higher reward to the positive path. For each query, we construct a pair of positive and negative paths via KG traversal. For medical and legal queries, positive paths are strictly grounded in logical supportive evidence, whereas negative paths do not follow the query logic. For each query in CWQ datasets, we leverage GPT-4o-mini to translate its SPARQL query statement into a positive path, and randomly replace entity or relation in the KG to generate negative paths. As shown in Table~\ref{tab:rm_result}, SCPRM consistently outperforms baselines across three datasets, which demonstrates its enhanced sensitivity to risky reasoning steps, particularly in the medical KGQA. 

\subsection{KG-based Inference Reward Comparison}

\begin{figure}
    \centering
    \includegraphics[width=\linewidth]{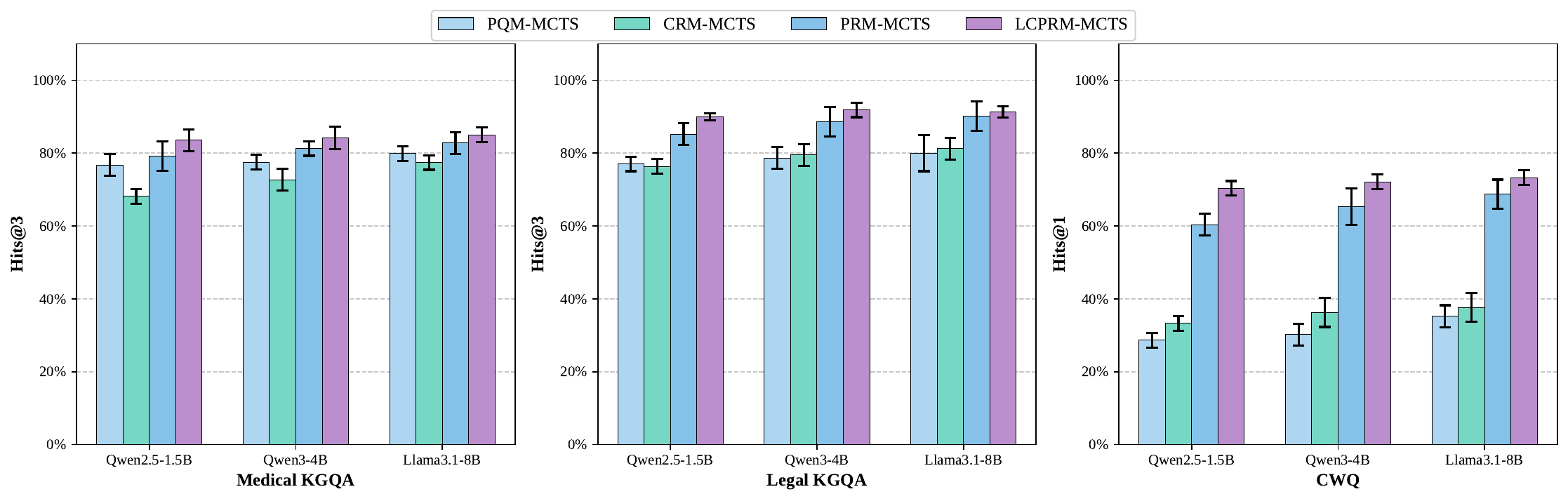}
    \caption{Hits@k performance (\%) of different reward models across datasets.}
    \label{fig:qa_infer}
\end{figure}

\begin{table}[htbp]
    \begin{minipage}[t]{0.5\linewidth}
        \small
        \caption{Hits@k performance (\%) of different methods across datasets. * indicates GPT-4o-mini, and ** denotes Llama3.1-8B.}
        \begin{tabular}{cccc}
        \toprule
        \multirow{2}{*}{Methods} & Medical & Legal & CWQ \\
        & Hits@3 & Hits@3 & Hits@1 \\
        \midrule
        ToG* & 83.06 & 80.69 & 70.87 \\
        ReKG-MCTS* & 82.47 & 81.26 & 72.78 \\
        SCPRM-MCTS** & 85.02 & 82.32 & 73.29 \\
        \midrule
        \end{tabular}
        \label{tab:rs_result}
    \end{minipage}
    \hfill
    \raisebox{-8pt}{
    \begin{minipage}[t]{0.45\linewidth}
        \vspace{0pt}
        To address \textbf{RQ2} and \textbf{RQ3}, we evaluate SCPRM on KG reasoning guided by MCTS, comparing against two distinct categories of recent KGQA methods. The first category consists of trainable reward model baselines with MCTS, where trainable baselines are integrated with same base LLMs, and the second category are training-free prompting methods based on LLM APIs.
    \end{minipage}
    }
\end{table}

The empirical results reveal several key insights. Firstly, within the trainable reward model paradigm, SCPRM-MCTS strictly outperforms all baseline reward models (PQM, CRM and PRM) across all three datasets shown in Figure~\ref{fig:qa_infer}. Secondly, despite relying on a 8B base LLM, our framework outperforms the GPT-4o-mini powered baselines (ToG and ReKG-MCTS) on the those datasets demonstrated in Table~\ref{tab:rs_result}. 

\subsection{Ablation Study}
To investigate the individual contributions of each component in SCPRM, we conduct an ablation study with two variants based on Llama3.1-8B, which are w/o CR (without cumulative reward) and w/o FR (without future reward). As Table~\ref{tab:ab_study} shows, removing CR causes severe degradation, with pairwise path ranking accuracy dropping by 9.85\% to 11.82\%, and Hits@k falling by 9.46\% to 12.15\% across datasets. Without CR, the model ignores the accumulated risk along the reasoning chain, leading to compounding errors. Notably, analyzing the metric differences reveals the distinct roles of the two components. While removing CR generally inflicts the most damage on step-level Accuracy, removing FR exhibits a remarkably acute impact on end-to-end Hits@k performance. For instance, in Legal KGQA, w/o FR causes a dramatic 12.30\% drop in Hits@3, even underperforming the w/o CR variant. This divergence indicates that CR is essential for establishing a reliable foundation by penalizing historical errors, whereas FR acts as a critical lookahead mechanism that refines path utility for final selection. Therefore, they provide orthogonal and synergistic reward signals together.
 
\begin{table}[htbp]
    \centering
    \small
    \caption{Performance (\%) of ablation study across datasets.}
    \begin{tabular}{ccccccc}
    \toprule
    \multirow{2}{*}{Methods} & \multicolumn{2}{c}{Medical KGQA} & \multicolumn{2}{c}{Legal KGQA} & \multicolumn{2}{c}{CWQ} \\
    & Accuracy & Hits@3 & Accuracy & Hits@3 & Accuracy & Hits@1 \\ 
    \midrule
    SCPRM & 95.34 $\pm$ 1.61 & 85.02 $\pm$ 1.89 & 88.11 $\pm$ 1.47 & 82.32 $\pm$ 2.24 & 81.87 $\pm$ 1.51 & 73.29 $\pm$ 3.27 \\
    w/o CR & 84.17 $\pm$ 2.78 & 72.87 $\pm$ 2.33 & 78.26 $\pm$ 3.48 & 72.56 $\pm$ 3.36 & 70.05 $\pm$ 3.97 & 61.87 $\pm$ 2.96 \\
    w/o FR & 89.73 $\pm$ 1.26 & 78.25 $\pm$ 3.09 & 80.34 $\pm$ 1.44 & 69.72 $\pm$ 2.49 & 71.26 $\pm$ 1.27 & 62.46 $\pm$ 3.54\\
    \bottomrule
    \end{tabular}
    \label{tab:ab_study}
\end{table}

\subsection{Robustness Analysis}
In practice, errors in parsing query schema and constructing reasoning schema for a given trajectory are practically unavoidable, inevitably introducing noise during the reward model training phase. For instance, an entity might possess multiple semantic types within a KG like Freebase, or a constructed reasoning schema may inadvertently violate valid path directions. To systematically investigate the resilience of SCPRM against such schema noise, we artificially inject varying proportions of noisy schema (ranging from 0.1 to 0.5) during the training phase and evaluate its impact on KG reasoning path reward estimation with MCTS.

\begin{figure}
    \centering
    \includegraphics[width=\linewidth]{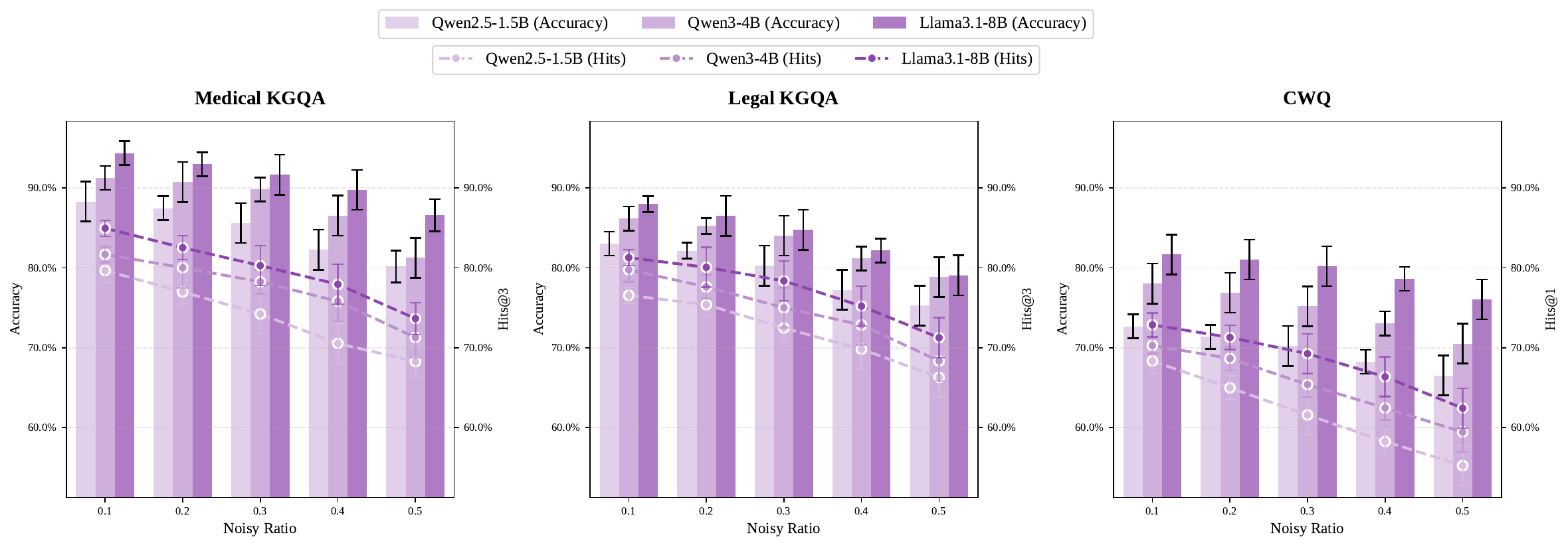}
    \caption{Performance (\%) of different noisy ratios of SCPRM across datasets.}
    \label{fig:robust_results}
\end{figure}

As illustrated in Figure~\ref{fig:robust_results}, we observe a consistent and intuitive trend across three datasets: while both pairwise path ranking accuracy and KG reasoning hits experience a gradual decline as the noise ratio increases, the degradation remains remarkably graceful. Notably, even under severe noise conditions (e.g., a 0.5 noise ratio), the models retain a robust level of performance without catastrophic failure. Furthermore, the empirical results indicate that larger base models (e.g., Llama3.1-8B) not only yield superior absolute performance but also exhibit enhanced robustness against noise compared to their smaller counterparts (e.g., Qwen2.5-1.5B). Overall, these findings substantiate the inherent robustness of SCPRM, demonstrating its capacity to effectively distill reliable reward signals even from highly noisy trajectories.

\section{Conclusion}
In this work, we introduce an $A^*$-style reward function that models both cumulative past rewards and future rewards for KG reasoning. By aggregating step-wise safety signals, the cumulative past reward formulates the existing trajectory quality. Meanwhile, the future reward estimate the future success, whose optimization align the reasoning guided by the implicit targets with the query and reasoning schema difference. The two components yield a principled and effective reward signal and improve reasoning performance.

\bibliographystyle{abbrv}
\bibliography{reference.bib}

\end{document}